\documentclass[letterpaper, 10 pt, conference]{ieeeconf}  

\IEEEoverridecommandlockouts                              

\usepackage{graphicx}
\usepackage{xcolor}
\usepackage{colortbl}
\usepackage{booktabs}
\usepackage{flushend}
\usepackage{wrapfig}
\usepackage{amsmath,amsfonts,amssymb}
\usepackage{ulem}
\usepackage{url}

\title{\LARGE \bf
A System for Imitation Learning of\\Contact-Rich Bimanual Manipulation Policies  
}

\author{Simon Stepputtis$^{1}$, Maryam Bandari$^{2}$, Stefan Schaal$^{2}$, Heni Ben Amor$^{3}$
\thanks{$^{1}$Simon Stepputtis is with the Robotics Institute, Carnegie Mellon University, Pittsburgh, United States
        {\tt\small stepputtis@cmu.edu}}%
\thanks{$^{2}$Maryam Bandari and Stefan Schaal are with [Google] Intrinsic, Mountain View, United States
        {\tt\small \{sschaal, maryamb\}@google.com}}%
\thanks{$^{3}$Heni Ben Amor is with the School of Computing and Augmented Intelligence, Arizona State University, Tempe, United States
        {\tt\small hbenamor@asu.com}}%
}

\usepackage{amsmath,amsfonts,bm}

\newcommand{\mb}[1]{\mathbf{#1}}

\def\eqref#1{equation~\ref{#1}}

\def\1{\bm{1}}

\def\vf{{\bm{f}}}

\def\vp{{\bm{p}}}

\def\mD{{\bm{D}}}

\def\mK{{\bm{K}}}

\def\mM{{\bm{M}}}

\def\mY{{\bm{Y}}}

\DeclareMathAlphabet{\mathsfit}{\encodingdefault}{\sfdefault}{m}{sl}
\SetMathAlphabet{\mathsfit}{bold}{\encodingdefault}{\sfdefault}{bx}{n}

\def\sR{{\mathbb{R}}}

\definecolor{LightCyan}{rgb}{0.88,1,1}
\usepackage{array,etoolbox}
\preto\tabular{\setcounter{magicrownumbers}{0}}
\newcounter{magicrownumbers}
\newcommand\rownumber{\stepcounter{magicrownumbers}\arabic{magicrownumbers}}

\begin{document}
\normalem

\maketitle
\thispagestyle{empty}
\pagestyle{empty}

\begin{abstract}
In this paper, we discuss a framework for teaching bimanual manipulation tasks by imitation. To this end, we present a system and algorithms for learning compliant and contact-rich robot behavior from human demonstrations. The presented system combines insights from admittance control and machine learning to extract control policies that can (a) recover from and adapt to a variety of disturbances in time and space, while also (b) effectively leveraging physical contact with the environment. We demonstrate the effectiveness of our approach using a real-world insertion task involving multiple simultaneous contacts between a manipulated object and insertion pegs. We also investigate efficient means of collecting training data for such bimanual settings. To this end, we conduct a human-subject study and analyze the effort and mental demand as reported by the users. Our experiments show that, while harder to provide, the additional force/torque information available in teleoperated demonstrations is crucial for phase estimation and task success. Ultimately, force/torque data substantially improves manipulation robustness, resulting in a 90\% success rate in a multipoint insertion task. Code and videos can be found at
\url{https://bimanualmanipulation.com/}
\end{abstract}

\section{Introduction}

\noindent
Manipulation still remains a critical challenge of robotics~\cite{billard2019}. Over the past decades, there has been tremendous progress in endowing robots with motor skills for grasping and dexterity. However, the vast majority of work in this field focuses on scenarios involving a single robot arm and tightly controlled physical interactions with the environment. With decreasing prices, as well as the proliferation of collaborative and humanoid robotics, there is increased need for techniques that enable reliable, efficient and safe bimanual manipulation. Bimanual robots need to perform manipulation tasks that involve multiple points of contact and dynamic force exchange with objects (and humans) in their surroundings, e.g., lifting a box, inserting a tight-fitting part, unscrewing a bottle cap, or reacting to a human push. However, making early or premature contact with a target object may create forces that seriously jeopardize the manipulation process. In addition to physical interaction with their surroundings, the individual robot arms in a bimanual setup may themselves be exchanging forces and torques through manipulated objects. These forces may lead to oscillations, instabilities and damage to the underlying hardware. Consequently, compliant control policies are required that allow bimanual robots to (a) deal and adapt to a wide variety of disturbances in space and time, and (b) effectively leverage physical contact to their advantage. Yet, designing control frameworks that can bridge these (potentially conflicting) requirements can be challenging and time-consuming. To date, only few publications have addressed such challenges underlying bimanual manipulation~\cite{kroemer2021review}.

\begin{figure}[t!]
    
    \centering
    \includegraphics[width=1\linewidth]{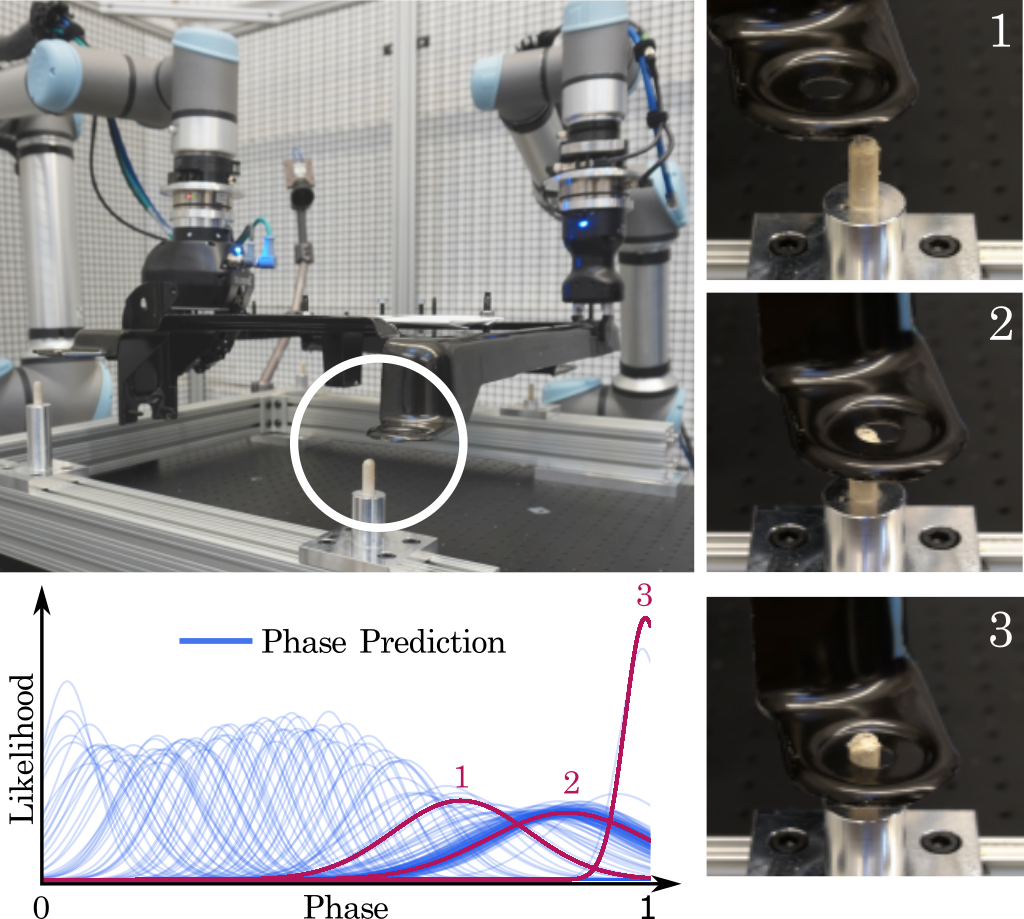}
    \caption{Overview of the bimanual insertion task in which the two robots need to jointly insert the bracket onto four pins. The phase plots and detailed pictures show how our method is capable of adjusting the bracket's position.\label{fig:teaser}}
\end{figure}

In this paper, we present a system for imitation learning of compliant bimanual manipulation policies. We describe a robotic setup in which human demonstrations of the task are recorded across a variety of sensing modalities. The setup leverages principles of admittance control to enable contact-rich and dynamic demonstrations without the risk of collisions, damage or wear-and-tear. In turn, the recorded data is used to learn Interaction Primitives which encode the demonstrated behavior in time and space. At runtime, interaction primitives are used to identify the temporal task progress as a function of external perturbations, as well as the optimal robot response to these perturbations and environmental conditions. In cases where  physical perturbations affect task execution, the robot is able to account for it by performing corrective actions in either time or space. Since the presented approach is Bayesian in nature, it allows for powerful spatio-temporal inference from multimodal datastreams.

\begin{figure*}[t!]
    \centering
    \includegraphics[width=\linewidth]{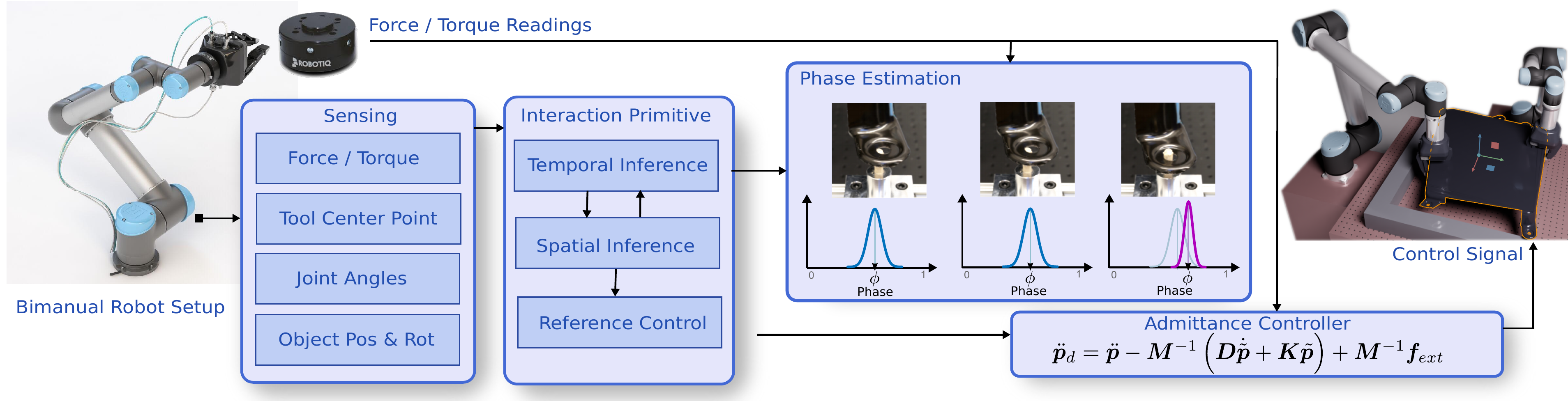}
    \caption{Overview of the proposed system: sensing data acquired from a number of multimodal sensors is used to learn a Bayesian Interaction Primitive. In case of misalignment, temporal phase estimation and its progression is used with an admittance controller to overcome obstruction. 
    }
    \label{fig:overview}
\end{figure*}

The main objective of this paper is to provide insights and solutions regarding a number of technical and theoretical considerations that have to be taken into account for contact-rich manipulation tasks in bimanual setups. We argue that four key components influence the performance of a system for imitation learning in bimanual manipulation. Specifically, we will discuss methodologies for a) \emph{data collection}, b) \emph{motor skill learning}, c) \emph{task phase estimation}, and  d) \emph{compliance through sensing and control}. A critical conclusion in this regard is the \textbf{importance of task phase estimation and phase monitoring} during behavior execution. 
\section{Related Work}

Bimanual manipulation requires the accurate coordination of multiple robot manipulators in the same real-time cycle. Early approaches largely focused on planning techniques in order to generate kinematic configurations that are feasible for both involved  arms~\cite{koga1992}. Due to the computational costs, more recent work~\cite{vahrenkamp2009humanoid} introduced precomputed reachability analysis and medial axis transforms to efficiently generate candidate robot configurations. In a similar vein, the work in~\cite{cuong2018} presents a certifiably-complete manipulation planner for assembly tasks involving two robots.
However, such planning-based approaches assume quasi-static motions and do not provide the responsiveness needed for contact-rich, dynamic manipulation tasks. An alternative approach is to use machine learning techniques in order to extract reactive manipulation policies. The work in~\cite{luck2017} used reinforcement learning (RL) in a low-dimensional space of synergies in order to find policies that optimally coordinate both arms. Amadio et al.~\cite{amadio2019} leverage symmetries in the kinematic structure to reduce the sample complexity of the trial-and-error process underlying reinforcement learning. Despite these optimizations, RL is a time-consuming process that does not leverage existing human knowledge about such bimanual tasks as found in industrial applications. In addition, RL on real robot platforms comes with a substantial burden of repeatedly resetting the experiment to its initial state, as well as wear-and-tear on robots and sensors. By contrast, imitation learning promises to use only a limited set of human expert demonstrations in order to extract an underlying policy. The work in~\cite{lioutikov2016learning} uses such an approach to learn primitives for bimanual manipulation. However, the work largely focuses on how to sequence individual primitives in order to realize a long-horizon task. Our work is closest related to the work in~\cite{pastor2011} and~\cite{ijspeert2013dynamical} in which Dynamic Motor Primitives (DMP) are used to learn bimanual manipulation policies from demonstrations. For an excellent overview article on robot learning for manipulation, we refer the reader to~\cite{kroemer2021review}.  
\section{Methodology}
In this section, we will introduce our system for imitation learning in bimanual settings and discuss a variety of considerations regarding data collection, learning and compliance. The system learns compliant bimanual manipulation policies from human demonstrations. Without loss of generality, we will  focus on a specific task wherein two robot arms are required to first lift a bracket, which features four alignment grommets. Once lifted, the bracket is to be carefully inserted onto a set four pegs via the four grommets. Due to the multiple distributed positions, task execution typically involves substantial physical contact between the bracket and the pegs. Accordingly, after first contact, the bracket position and orientation may have to be repeatedly corrected for successful insertion. For an overview of this task see Fig.~\ref{fig:teaser}.

\subsection{Data Collection for Bimanual Manipulation} 
\begin{figure}[t!]
    \centering
    \includegraphics[width=0.95\linewidth]{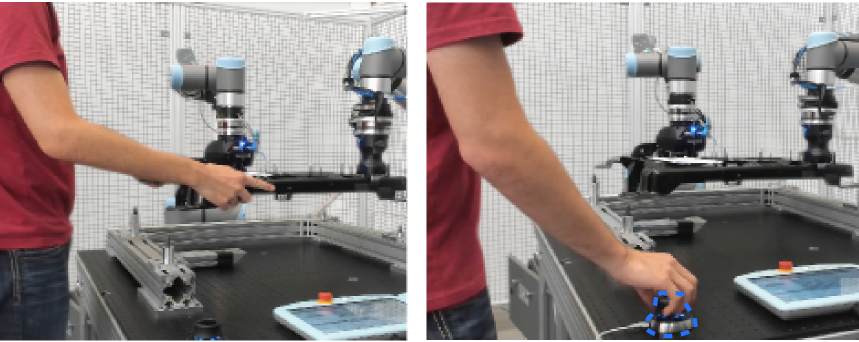}
    \caption{Data Collection for imitation learning. Left: direct, physical contact allows kinesthetic teaching. Right, the robot arms are controlled via a space mouse. In the above example, the user controls the tool center point of the two robots.}
    \label{fig:datacollection}
\end{figure}

An important first consideration when learning such a delicate manipulation task is the collection of training data. Our system provides two alternative approaches to data collection, namely kinesthetic teaching~\cite{billard2008handbook} and tele-operation, as can be seen in Fig.~\ref{fig:datacollection}. Kinesthetic teaching allows the human expert to provide demonstrations through physical guidance. In the bimanual setup, this can be achieved by either directly touching the involved robots or by moving the manipulated object, thereby applying forces on the attached hands of the robot. Alternatively, a \emph{Space-Mouse} can be used for data collection. A Space-Mouse is a 6 degree-of-freedom (DoF) input device that was first developed for the control and teleoperation of robot arms in space, in particular for the Robot Technology Experiment on the Spacelab D2 mission~\cite{hirzinger2004}. Accordingly, its design and functionality is optimized for the demands of manipulation tasks. As can be seen in Fig.~\ref{fig:overview}, a number of sensing modalities are continuously recorded during training, i.e., force-torque values, joint angle readings, tool center points, as well as the position and orientation of the manipulated object. 

\subsection{Motor Skill Learning and Temporal Inference} Given the recorded set of demonstrations, a key next step is to extract a policy or motor skill that generalizes the observed behavior to new situations. A variety of methods can be used for this purpose. Behavioral cloning (BC) with neural networks~\cite{pomerleau1991} and Probabilistic Motor Primitives (ProMP)~\cite{paraschos2013probabilistic} are among the most prominent methods. However, the above methods purely focus on the spatial aspects of motor control, as do most other techniques for imitation learning of motor skills. As a result, robots are not empowered to reason about the temporal evolution of a task and how time progresses. However, for successful bimanual manipulation in contact-rich tasks, robots need to constantly monitor and reevaluate the temporal progress of task execution. As a result of force interactions between the robot and the environment, as well as between the individual manipulators, motor commands may not get accurately executed. Such a situation may be due to a variety of conditions such as physical obstructions, friction, an externally applied force, etc. To overcome such bottleneck situations, it is critical that the robot generates motor commands that are temporally-adequate until all obstacles are overcome. 

\emph{Temporal Inference:} In our framework, we adopt a methodology for temporal reasoning inspired by work on human-robot interaction (HRI)~\cite{amor2014interaction}. In HRI scenarios, robots need to constantly re-estimate the current temporal phase rather than relying on a predefined internal clock for its progression. In this context, the term \emph{phase} describes the relative temporal position within a task, i.e., the phase variable may be zero at the beginning of a task and one when the task is finished. In HRI, phase estimation is performed by observing the human partner's movements and, in turn, inferring the most likely position along the time dimension. The work in~\cite{campbell2017bayesian} showed that this procedure is akin to performing robot localization in time rather than space. Bayesian Interaction Primitives (BIP) ~\cite{campbell2019probabilistic} is an imitation learning approach for HRI which leverages this insight to perform both spatial and temporal reasoning. 

In our system, we use BIP in realtime to infer the spatial and temporal state of the execution of the manipulation task. However, rather than estimating the phase by observing an external, human partner, we use the multimodal sensing sources on the robot itself, e.g., force-torque and joint angles. An example of this process can be seen in Fig.~\ref{fig:overview}. In this example, a bracket is carefully placed on four pegs. In the first two images, the object appears stuck on-top of the pegs and insertion cannot proceed successfully. In the plots below the image sequence, we see the estimated phase along with the corresponding variance (i.e. the uncertainty in estimation) visualized as a Gaussian distribution. We notice that the phase estimate roughly remains constant. In the final image, we see that the physical obstruction has been overcome. Accordingly, the phase estimate now moves forward in time. This ability to estimate the phase allows the robot to carefully monitor the progress of a task so as to determine whether to continue task execution or to perform a refinement action. Note that in a BIP the spatial and temporal inference go hand in hand -- the robot determines both \emph{what to do} and \emph{when to do it}.

\subsection{Bayesian Interaction Primitives}
Our overall goal in this paper is to learn a bimanual manipulation policy that can generate accurate robot controls from observed states. Each recorded demonstration $\mY \in \mathbb{R}^{D \times T}$ contains $T$ observations with a $D$-dimensional feature vector.  Given the current state of the robot, the policy should generate optimal control signals that mimic the behavior of the human demonstrator. However, in the context of human-robot interaction and physical contact with the environment, it is also critical that the manipulation policy adapts to external perturbations. 
This may be, for example, a push from a human partner or forces generated from premature contact with the environment.

To learn such a policy, we use Ensemble Bayesian Interaction Primitives (EnBIP)~\cite{campbell2019probabilistic} (\footnote{Code available at \url{https://github.com/ir-lab/intprim}}) and learn a generative probabilistic model over the training demonstrations. However, instead of immediately using time-discretized data, EnBIP transform the recorded demonstrations into a time-invariant representation by performing a basis function decomposition. Such decompositions have been popularized in~\cite{ijspeert2013dynamical}, and have since been used in a number of motor primitive formulations~\cite{paraschos2013probabilistic, amor2014interaction}. Applying the basis function decomposition, we now approximate each state dimension $\mb{Y}^d_{t} = \mb{\Phi}_{\phi(t)}^{\intercal} \mb{w}^d + \epsilon_y$ through a linear combination
of $B$ basis functions $\mb{\Phi}_{\phi(t)} \in \mathbb{R}^{B^d}$ with corresponding basis weights $\mb{w}^d \in \mathbb{R}^{B^d}$. Note the shift into a relative time measure known as \emph{phase} $\phi(t) \in \mathbb{R}$, where $0 \leq \phi(t) \leq 1$, as well as the approximation error $\epsilon_y$. Intuitively, this allows the model to represent demonstrations with different lengths with the virtual length $\phi(t)$. 
The above decomposition generates $d$ weight vectors $\mb{w}$ of the same length $B$ and is therefore a practical first step towards efficient encoding and modelling of the recorded data in a compact fashion. Concatenating all basis weight vectors for the individual dimensions then forms the compressed representation $\mb{w} = [\mb{w}^{0\intercal}, \dots, \mb{w}^{D\intercal}] \in \mathbb{R}^{B}$ of a given trajectory where $B = \sum_d^D B^d$. In turn, we can now extract a probability distribution $p(\mb{w})$ over all given demonstrations. Sampling from this distribution generates a sample trajectory containing all observed sensor states and robot controls in the bimanual task. Similarly, we can also condition on previous states of the robot to yield a posterior distribution $p(\mb{w}_t | \mb{Y}_{1:t}, \mb{w}_{0})$ over \emph{future} states and controls.

However, a critical insight in BIP is the interplay between temporal and spatial reasoning. An estimation error can be the result of either errors in time or space. So far, inference assumes that the current time step or phase are known. This is typically only the case, when the robot's behavior is unencumbered by physical interactions with humans or the environment. To enable such adaptation, BIP reformulates the problem as a joint spatio-temporal inference -- time and space are coupled and are jointly estimated. This insight is realized by forming a new state vector $\mb{s} = [\phi, \dot{\phi}, \mb{w}]$. The state vector holds both a spatial component, contained in the basis function weights $\mb{w}$, as well as temporal components in the form of the phase $\phi$ and the phase velocity $\dot{\phi}$. The temporal variables $\phi$ and $\dot{\phi}$ describe where we are in time and how fast we are progressing with the task. Hence, generating the posterior:
\begin{equation}
\label{eq:ip_general}
    p(\mb{s}_t | \mb{Y}_{1:t}, \mb{s}_{0}) \propto p(\mb{y}_{t} | \mb{s}_t) p(\mb{s}_t | \mb{Y}_{1:t-1}, \mb{s}_{0}).
\end{equation}
now yields both the information about the spatial and temporal aspects of robot control, since these are encoded in $\mb{s}_t$. Performing the above inference step can efficiently be done via recursive filtering. More specifically, we use an Ensemble Kalman Filter (EnKF) as proposed in~\cite{Roth2017}. A major advantage of EnKF is its ability to model complex nonlinear distributions without having to specify any parametric family. The initial distribution is immediately formed by the provided demonstrations -- no fitting to a parametric family of densities is needed. In addition, EnKF can be used with  nonlinear transition functions and observation functions. As a result, linearization errors as found in other filters can be avoided. When compared to particle filters (PF)~\cite{thrun2005probabilistic}, EnKF avoid the problem of sample degeneracy and are typically more sample-efficient, i.e., need less ensemble members, than PF.

We start by defining an ensemble $\mb{X}$ of $E$ members shown by $\mb{X} = [\mb{x}^1, \dots, \mb{x}^E]$.
Optimally we want to sample the initial ensemble $\mb{X}_0$ directly from the prior $\mb{x}_0 \sim p(\mb{w}_0)$ for all $\mb{x}_0 \in \mb{X}_0$; however, since we do not have direct access to $p(\mb{w}_0)$, as a data-driven method, it is standard to instead sample from observed training demonstrations.
Random selection on ensemble members is reasonable as the ensemble-based filtering approach provides robustness against possible non-Gausian uncertainties, provided the number of ensemble members is not less than the number of example demonstrations $E \leq N$.
As a two-step Bayesian estimation method, our first step approximates $p(\mb{w}_t | \mb{y}_{1:t-1}, \mb{w}_{0})$ by propagating each ensemble member forward one time step with:
\begin{align}
\label{eq:state_prediction}
\mb{x}^j_{t|t-1} &=
g(\mb{x}^j_{t-1|t-1})
+
\epsilon_x, \quad 1 \leq j \leq E,
\end{align}
with constant-velocity state transition operator $g(\cdot)$, and noise error $\epsilon_x$.
Next, the ensemble members are updated from the observation and the nonlinear observation operator $h(\cdot)$:
\begin{align}
\mb{H}_t\mb{X}_{t|t-1} &= \left[h(\mb{x}^1_{t|t-1}), \dots, h(\mb{x}^E_{t|t-1})\right]^\intercal, \\
\mb{H}_t\mb{A}_t &= \mb{H}_t\mb{X}_{t|t-1} \\
&-\left[ \frac{1}{E} \sum_{j=1}^{E}h(\mb{x}^j_{t|t-1}), \dots, \frac{1}{E} \sum_{j=1}^{E}h(\mb{x}^j_{t|t-1}) \right], \nonumber
\end{align}
The deviation of each ensemble member from the sample mean $\mb{H}_t\mb{A}_t$ and the observation noise matrix $\mb{R}$ can then be used to compute the innovation covariance with:
\begin{align}
\mb{w}_t &= \frac{1}{E - 1} (\mb{H}_t\mb{A}_t) (\mb{H}_t\mb{A}_t)^\intercal + \mb{R}.
\end{align}
The Kalman gain is likewise calculated directly from the ensemble, with no need to specify an explicit covariance matrix, with
\begin{align}
\mb{A}_t &= \mb{X}_{t|t-1} - \frac{1}{E} \sum_{j=1}^{E}\mb{x}^j_{t|t-1}, \\
\mb{K}_t &= \frac{1}{E - 1} \mb{A}_t (\mb{H}_t\mb{A}_t)^\intercal \mb{w}^{-1}_t.
\end{align}

As is typical in recursive filtering, partial observations are sufficient to optimally estimate the full state, which we leverage to generate a posterior over unobservable latent variables, i.e. robot controls. Since the posterior is over weights $\mb{w}$ it defines the controls for all future time steps. 

By performing this inference scheme in each time step, we can generate posterior distributions that are conditioned on a multitude of sensors. 
In Fig.~\ref{fig:real_world_perturbation}, for example, we can see the effect of conditioning an execution on high sensor readings from the force-torque sensor. In this specific case, the robot learned to change direction away from this force exchange. 

\begin{figure*}[]
    \centering
    \includegraphics[width=1\linewidth]{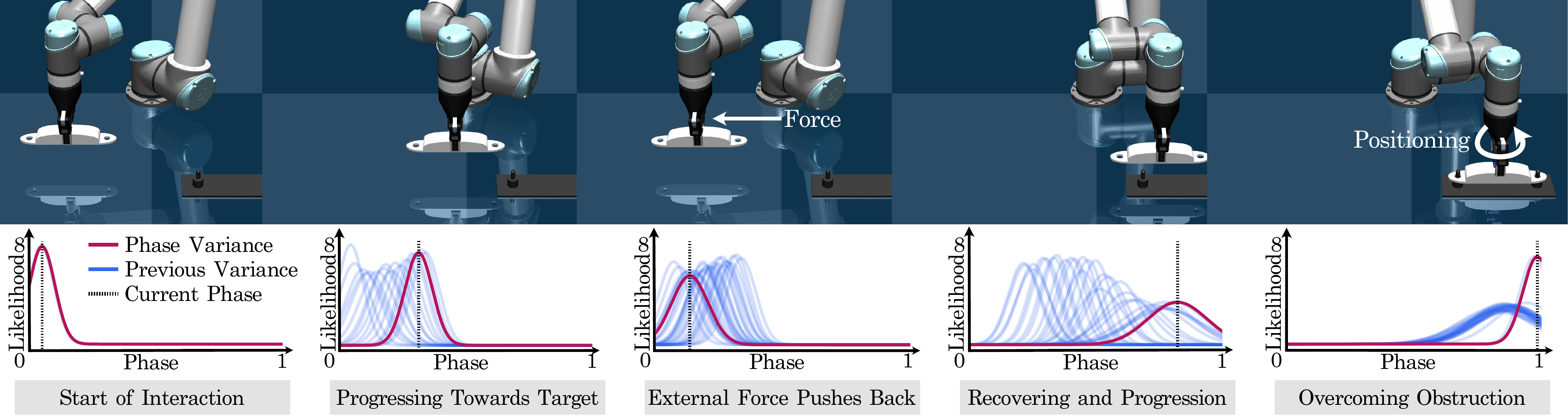}
     \caption{Image sequence showing the realtime phase estimation $\phi$ and its variance during the insertion of a bracket. The variance indicates the uncertainty underlying our estimates. In the third image, the robot is pushed backwards through a force that is applied to its gripper. In turn, the phase estimate is adjusted backwards in time during inference.
     In the final image, a physical obstruction causes the robot to get stuck (visible in the overlapping blue lines). Lastly, even the obstruction is overcome and the phase converges to value one, i.e., the goal is reached.}
     \label{fig:image_sequence_simulation}
\end{figure*}

\subsection{Admittance Control}

The last layer of our stacked control system is composed of a Cartesian admittance controller (suitable for a UR position controlled robot) with a variable target pose $\vp_d \in \sR^{6}$ and target velocity $\dot{\vp}_{d} \in \sR^{6}$. The pose $\vp$ is the concatenation of the robot's Cartesian position and rotation. Internally, rotations are computed with quaternions; however, for notational simplicity, we use Cartesian rotations. While the interaction primitive is performing inference over both robots jointly, the admittance controllers are running separately on each of the two robots with different hyper parameters. The use of admittance control is critical in order to avoid force accumulations due to the closed kinematic-chain, contact with the environment, or force interactions with a human user.

For simplicity, the following description of the controller describes the setup for a single robot. Fundamentally, the admittance controller implements the virtual dynamics calculating the external force and torque $\vf_{ext} \in \sR^{6}$ in Cartesian space.
\begin{equation}
    \label{eq:admittane_dynamics}
    \mM\Ddot{\tilde{\vp}} + \mD\dot{\tilde{\vp}} + \mK\tilde{\vp} = \vf_{ext}
\end{equation}
where the hyper-parameters are the virtual inertia $\mM \in \sR^{6\times 6}$, virtual Cartesian stiffness $\mK \in \sR^{6\times 6}$, virtual damping $\mD  \in \sR^{6\times 6}$, and pose error $\tilde{\vp} = \vp - \vp_d$ with current pose $\vp$. These parameters are set differently for each of the two robots to account for their individual dynamics. Given the desired control signal $\vp_{d}$ from the interaction primitive, target velocity $\dot{\vp}_{d}$ of zero, and external force and torque measure $\vf_{ext}$, the robots' control signal $\Ddot{\vp}_d$ is computed from equation~\ref{eq:admittane_dynamics} as follows
\begin{equation}
    \Ddot{\vp}_d = \Ddot{\vp} - \mM^{-1}\left(\mD\dot{\tilde{\vp}} + \mK\tilde{\vp}\right) + \mM^{-1} \vf_{ext}
\end{equation}
The resulting target position $\vp$ and velocity $\dot{\vp}$ is calculated through integration from $\Ddot{\vp}_d$. This calculation is done separately for each robot and is sent to be executed with an inverse kinematics to get the desired joint angles for each robot.

\section{Evaluation}
We evaluate our approach in two separate settings. In the first setting, a single, simulated robot is placing a bracket with two grommets onto two pins (see Fig.~\ref{fig:image_sequence_simulation}). In this setup, a single 6 Degree of Freedom (Dof) UR5 robot is equipped with a force-torque sensor between the flange of the arm and a parallel jaw gripper, tasked with placing one of two different brackets with varying tolerances (1mm and 5mm) onto two pins. The simulated environment is implemented in MuJoCo~\cite{todorov2012mujoco}. In a second, bimanual setup on two real robots (as previously introduced in Fig.~\ref{fig:teaser}), a UR5 robot and a UR10 robot with 6Dof each are used. The two robots are equipped with force-torque sensors located between the end-effector and the gripper. However, while the UR5 robot uses a parallel jaw gripper, the UR10 includes an adaptive three finger gripper for increased stability during the grasp, thereby preventing undesired tilting of the object. 

In both settings (single-arm in simulated and dual-arm in real experiments), we utilize a path planner with pre-determined way points to pick up the bracket from slightly randomized position to perform the initial lift while ensuring different grasping poses. The initial bracket position is varied within a margin of $\pm$1~cm. However, approaching the pins, as well as the final insertion task is performed via the described imitation learning model. Sufficiently precise manipulation is required to successfully insert the bracket given the 1 and 5~mm tolerance in simulation and 6~mm tolerance per pin in the real world bimanual setup. Note that this task does not cover Transition or Interference fits with potentially negative tolerances. In the simulated task, forces largely stem from contacts with surrounding objects while in the real-world task, forces may also be exchanged between the two robot arms during manipulation.

In the following sections, we evaluate the proposed system with respect to the different components, i.e., data collection, motor skill learning, as well as compliance and dynamic phase estimation. 
We also compare our method with a behavior cloning (BC) and ProMP~\cite{paraschos2013probabilistic} baseline. The latter is another common method derived from Dynamic Motor Primitives; however, it requires separate methods for spatio-tempora alignment of the motion.

\subsection{Data Collection}

To collect the required data for training the model, we use 30 demonstrations for the simulated and real-world setup each.  
For the simulated environment, 30 demonstrations are collected from pre-programmed and slightly randomized B\'ezier curves that alter the curvature of the generated motion, utilizing a 1mm tolerance bracket. Using the bracket with the smaller tolerance for training purposes is expected to also yield a successful model for the bracket with larger tolerances.

Similar to the simulated setup, 30 demonstrations are collected in the real-world setup. Fundamentally, demonstrations are collected from eight human subjects for a total of four datasets, utilizing either the Space-Mouse, or kinesthetic teaching. Additionally, one dataset each is collected with varying starting positions of the bracket within a margin of three centimeters. 
With the Space-Mouse, the participants control the position of the bracket, inherently moving the two robots via the reference poses $\vp_{ref}$ of each robot's admittance controller through a fixed transformation from the bracket's pose. When using kinesthetic teaching, the underlying admittance controllers allow the participants to freely move the bracket inside the overlapping workspace of the robots by continuously updating the controller's reference pose with the bracket's sensed position. 
To familiarize the participants with the intricacies of each training method, each participant had the opportunity to provide an initial, unrecorded demonstration. 

\subsubsection{Kinesthetic vs. Space-Mouse Data}
A fundamental difference between the training methods is the availability of force-torque sensor data. When using kinesthetic teaching, external forces are induced by the human teacher during the demonstration, rendering sensed force-torque data unusable for subsequent skill learning. In contrast, collected force-torque data when using the Space-Mouse can be used for skill learning as only contact forces with the environment and forces resulting from the interaction of the two robots are measured. 

\begin{figure}[t!]
    \centering
    \includegraphics[width=1\linewidth]{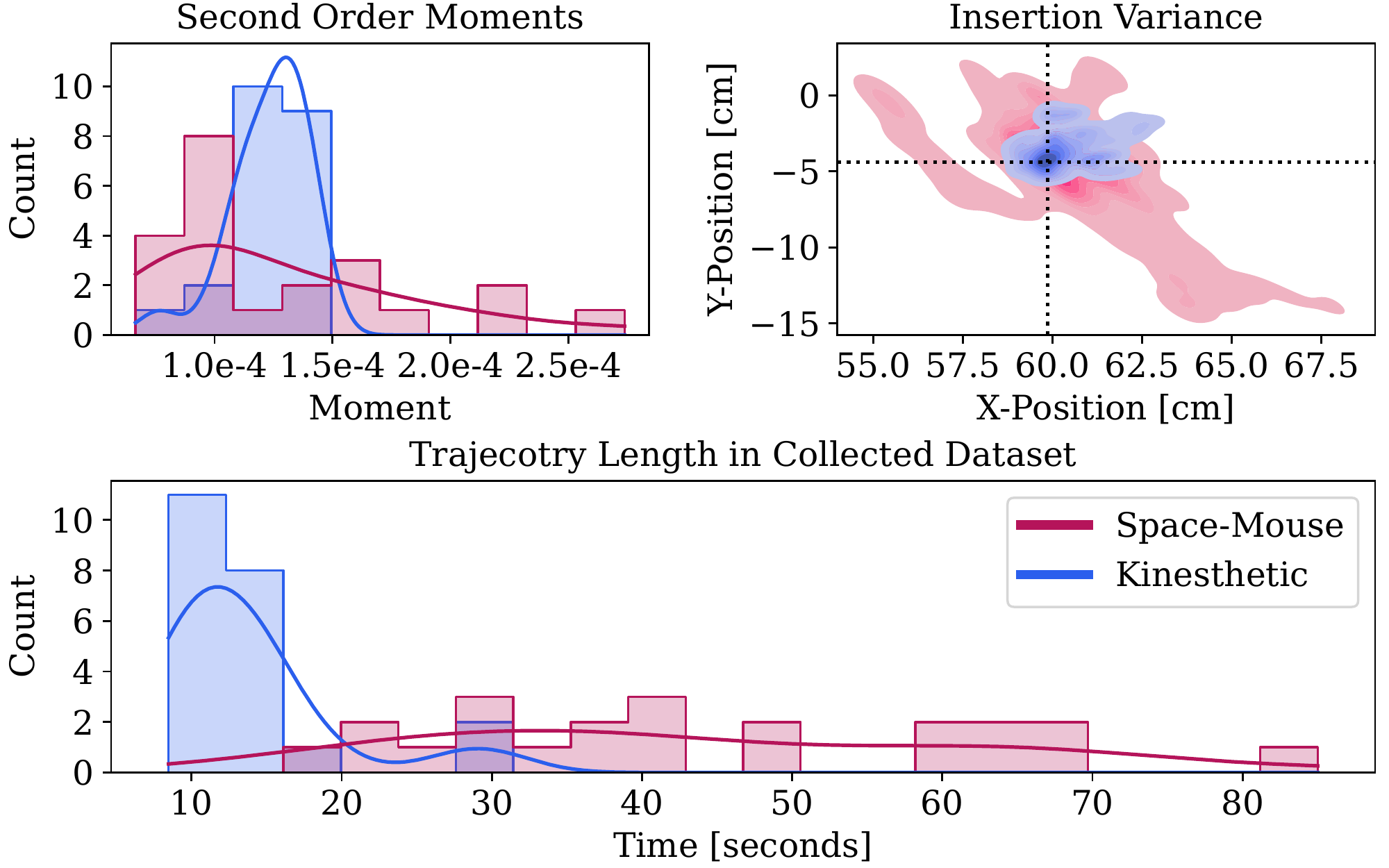}
     \caption{Analysis of the recorded user data with regards to smoothness (top left), variance close to the final insertion (top right) and demonstration length (bottom).}
     \label{fig:user_analysis}
\end{figure}

Figure~\ref{fig:user_analysis} shows a comparison of the demonstrations collected with kinesthetic teaching (blue) and the Space-Mouse (red) on the real robot. Generally demonstrations collected by using kinesthetic teaching result in smoother motion, shorter trajectories and reduced final adjustments prior to the final insertion. Prior to the final insertion of the pins, demonstrations collected with the Space-Mouse have a variance of 2.173~cm of the bracket location, while demonstrations collected with kinesthetic teaching show a significantly lower variance 0.886~cm. The need for final adjustments is also reflected in the average trajectory length of 44 and 14 seconds for the Space-Mouse and kinesthetic method, respectively. 

\subsubsection{NASA TLX Workload}

In addition to evaluating the collected data itself, we also evaluated how the human participants perceive the workload of providing demonstrations with each teaching method. While \textit{workload} is a subjective measure and varies between different participants, the NASA Task Load Index~\cite{nasa_tlx,nasa_tlx_original} provides a methodical assessment of the workload a user is experiencing when completing a task. We evaluate six categories: mental, physical, and temporal demand, as well as perceived performance, required effort and the users' frustration using each of the training modalities. 
However, since every user has a different subjective assessment of how much each category influences the overall perceived workload, weights for each category are derived prior to the assessment to further increase the sensitivity of the metrics across multiple subjects.

\begin{figure}[t!]
    \centering
    \includegraphics[width=1\linewidth]{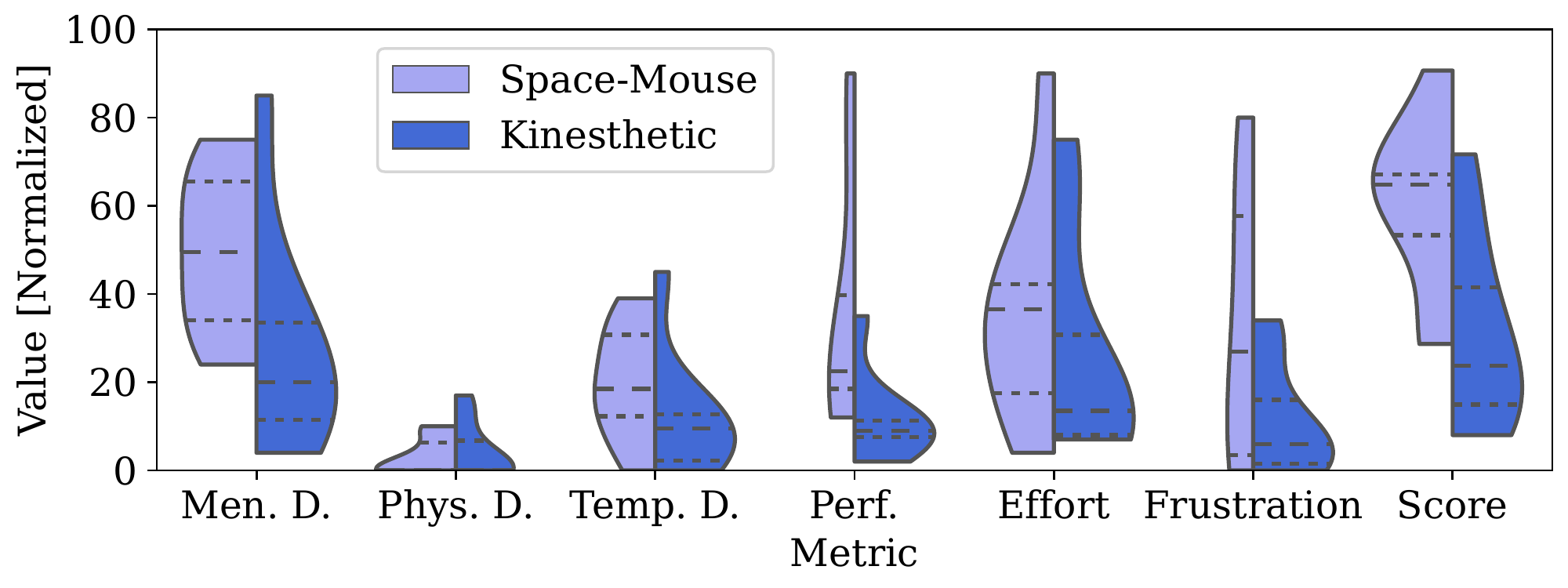}
     \caption{Nasa TLX Workload evaluation. Metrics are weighted by how important the participants' deemed them and normalized to [0, 100]}
     \label{fig:tlx}
\end{figure}

Figure~\ref{fig:tlx} shows the results of the workload assessment across eight users. Particularly the perceived performance as well as the frustration of the participants are significantly increased when using the Space-Mouse. This results is an interesting dichotomy: despite higher frustration, successfully completing the task with the Space-Mouse significantly increased the participants' perception of performance.
Overall, participants rated using the Space-Mouse as approximately twice as labour-intensive as kinesthetic teaching, with means of 59.33 and 30.88 respectively.  

While the analysis of the data as well as the perceived workload from the users, positions kinesthetic teaching as favorable, its major drawback, however, is the inability to record meaningful force-torque sensor data.

\subsection{Motor Skill Learning}

\begin{table}[]
\scriptsize
\setlength{\tabcolsep}{1.1em}%
\begin{tabular*}{\columnwidth}{@{\makebox[1em][r]{\rownumber\space}} @{\extracolsep{\fill}} c >{\columncolor[gray]{0.925}}c >{\columncolor[gray]{0.925}}c >{\columncolor[gray]{0.925}}c c}
\toprule
\multicolumn{1}{c}{} & \multicolumn{3}{c}{\cellcolor{gray!18} Dataset Variations} & \\
\multicolumn{1}{c}{Collection Method} & Extended  & With F/T Data  & Perturbed & Success  \\ \midrule
Space-Mouse &            &            &            & 100.0\% \\
Space-Mouse &            &            & \checkmark & 13.33\% \\
Space-Mouse &            & \checkmark &            & 90.0\% \\
Space-Mouse &            & \checkmark & \checkmark & 20.0\% \\
Space-Mouse & \checkmark &            &            & 100.0\% \\
Space-Mouse & \checkmark &            & \checkmark & 70.0\% \\
Space-Mouse & \checkmark & \checkmark &            & 93.3\% \\
Space-Mouse & \checkmark & \checkmark & \checkmark & 90.0\% \\ \midrule

Kinesthetic &            &            &            & 100.0\% \\
Kinesthetic &            &            & \checkmark & 0.0\% \\
Kinesthetic & \checkmark &            &            & 100.0\% \\ 
Kinesthetic & \checkmark &            & \checkmark & 73.33\% \\ \bottomrule

\end{tabular*}
\caption{Evaluation results on the rea-world bimanual setup 
}
 \label{tab:results}
\end{table}

\setcounter{magicrownumbers}{0}
\begin{table}[]
\scriptsize
\setlength{\tabcolsep}{0.2em}%
\begin{tabular*}{\columnwidth}{@{\makebox[1em][r]{\rownumber\space}} @{\extracolsep{\fill}} c ccc}
\toprule
\multicolumn{1}{c}{\textbf{Model}} & Tolerance & Disturbance & Success  \\ \midrule
ProMP & 5mm &            & 100.0\% \\
ProMP & 5mm & \checkmark & 0.0\% \\
ProMP & 1mm &            & 33.33\% \\ \midrule

EnBIP   & 5mm &            & 100.0\% \\
EnBIP   & 5mm & \checkmark & 100.0\% \\
EnBIP   & 1mm &            & 90.0\% \\ \bottomrule

\end{tabular*}
\caption{Results in simulated experiments }
 \label{tab:simulation}
\end{table}

Table~\ref{tab:results} shows the results of EnBIP on the real-world bimanual insertion task. We evaluated our approach with demonstrations accounting for varying starting positions (column \textit{Extended Dataset}), usage of force-torque data (column \textit{Force Sensor}), and variable starting positions of the bracket (column \textit{Varied Position}). 
A test is counted as successful if all four grommets are placed on the correct pin and both robots released the bracket. The success rate is reported over 30 evaluations. 

Using either dataset, a successful motor skill can be learned when the bracket is in a fixed starting position, resulting in a 100\% success rate (lines 1, 5, 9, and 11 in Table~\ref{tab:results}), even when the datasets are extended via demonstrations with varying starting positions of the bracket. However, when varying the starting position of the bracket during testing, we observe a 30\% drop (lines 6 and 12) in success rate. This can be attributed to the robots not being able to accurately sense the obstacle when attempting final insertion. When adding the force-torque sensor data, which is exclusively available in the dataset collected with the Space-Mouse, the success rate increases to 90\%, (line 8 of Table~\ref{tab:results}). Adding force-torque sensor data is therefore critical for robustness when varying starting positions of the bracket and results in an overall performance improvement of 20\%, compared to a setting in which no force-torque data is used (line 6 compared to line 8). As a result, while kinesthetic teaching produces cleaner data and is easier for participants to perform, the availability of force-torque data dramatically increases robustness in contact-rich tasks. 

\subsection{Dynamic Phase Estimation}

\begin{wrapfigure}{r}{0.5\linewidth}
        \includegraphics[width=0.99\linewidth]{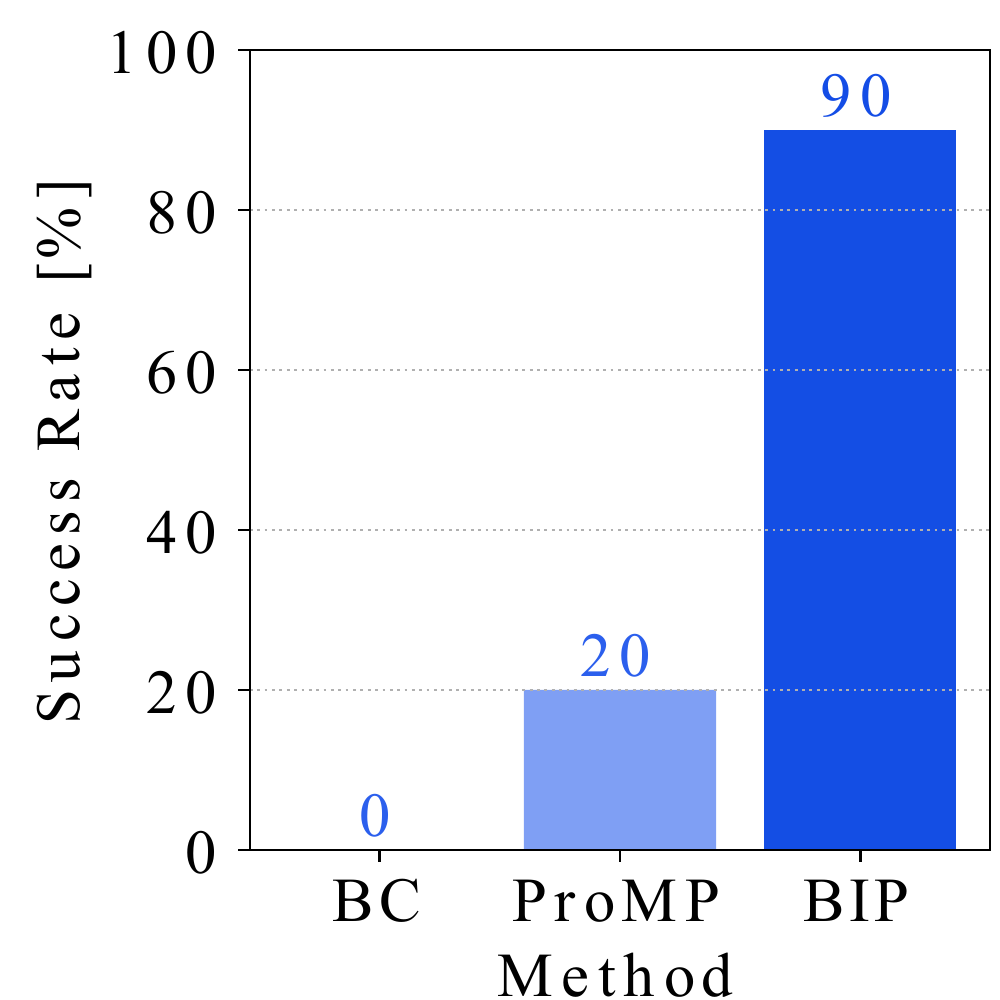}
     \caption{Baseline comparison with varying starting pose}
     \label{fig:baselines}
\end{wrapfigure}

To compare the impact of online phase estimation in the presence of external disturbances, we compare EnBIP to ProMP on two different brackets in the simulated environment. Figure~\ref{fig:image_sequence_simulation} shows the task in simulation on the top, as well as the current (red) and history of previous (blue) phase estimate.
Just after introducing an external force in picture two, our model only slightly increased its uncertainty during the reverse motion (step 3); however, in step five where the robot has placed the bracket on the pins, the uncertainty has increased just before the final insertion was made. A similar behavior to the latter can also be seen in Figure~\ref{fig:teaser} in the real-world experiment.
This example shows the crucial influence of phase estimation and correction for successful task completion.

\begin{figure}[]
    \centering
    \includegraphics[width=1\linewidth]{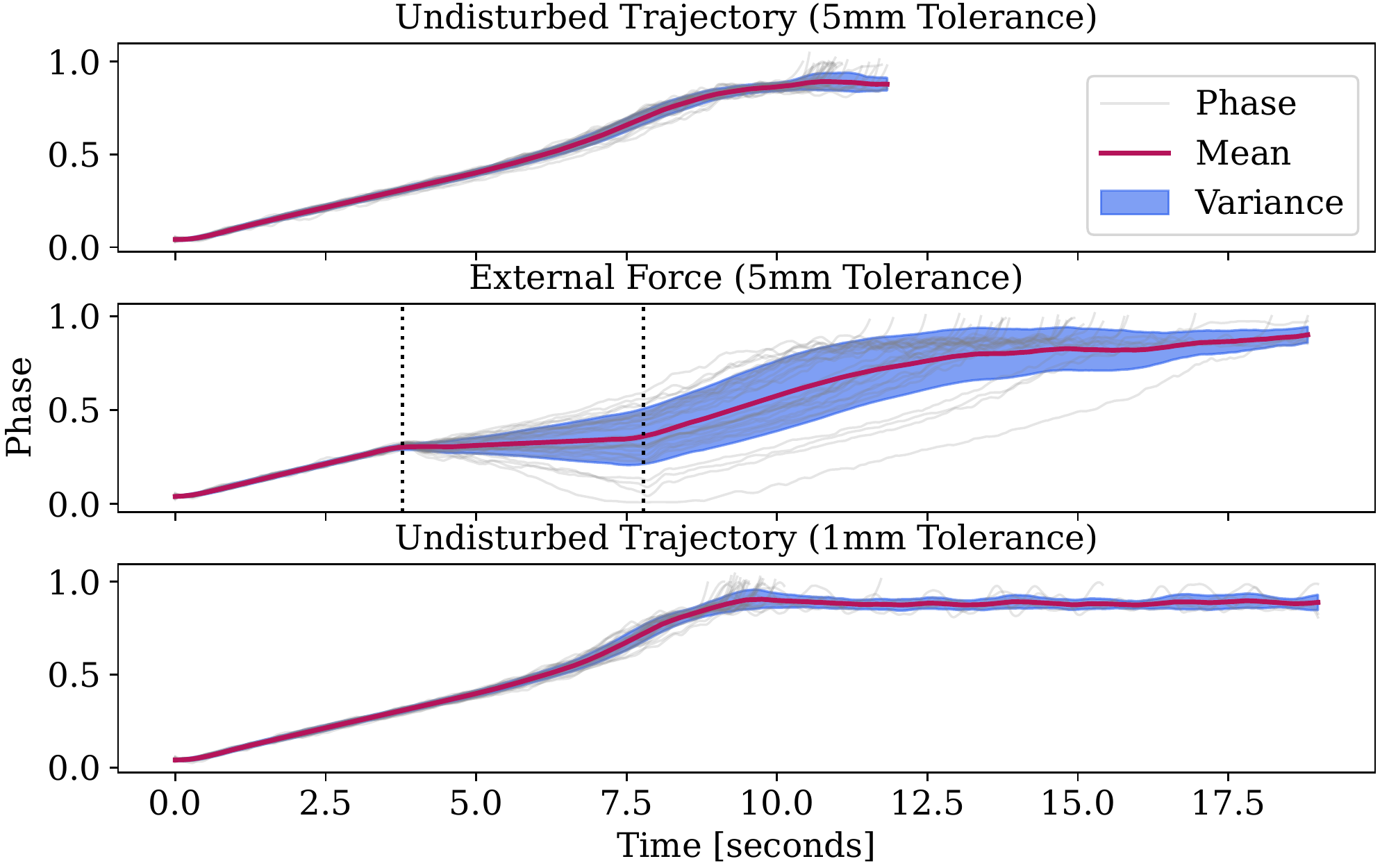}
     \caption{Phase progression in three scenarios: Undisturbed (top), External force (middle), and small tolerances (bottom) in the simulated environment.}
     \label{fig:conditioning_simulation}
\end{figure}

Figure~\ref{fig:conditioning_simulation} shows the estimated phase progression over 30 trials in the simulated environment of Figure~\ref{fig:image_sequence_simulation} for three different scenarios: At the top, the bracket with 5mm tolerance is placed on the two pins without any external disturbance, expecting a diagonal line from phase 0 to 1. The middle figure introduces an external force applied between 4.5 and 7.5 seconds. The applied force is drawn from a normal distribution with a mean of 1.75 Newton and a variance of 0.5 Newton. Gray lines show the actual phase estimation of the 30 trials in addition to the mean (red) and variance (blue). Even though the motion is disturbed, the model is able to gracefully recover by adjusting the phase, ultimately succeeded in $100\%$ if the tasks (Table~\ref{tab:simulation} line 4 and 5). Finally, the lowest plot shows the phase progression with the 1mm tolerance bracket without any disturbances, besides the ones occurring directly upon insertion of the pins. This bracket requires a higher precision, thus while a straight diagonal line from start to end would be expected, the model adjusts the bracket's position on the pins for an extended period of time. Ultimately the model succeeds in $90\%$ of the trials while failures are due to prematurely estimating task completion, or not finding the pins (Table~\ref{tab:simulation} line 4 and 5). 

Table~\ref{tab:simulation} shows a comparison with ProMP in the simulated environment, utilizing a fixed phase progression that is sampled uniformly across the lengths of the training demonstrations.
Line 5 shows a significantly improved success rates over ProMP in line 2 when external forces are applied to the system, underlining the importance of phase estimation.

\begin{figure}[]
    \centering
    \includegraphics[width=1\linewidth]{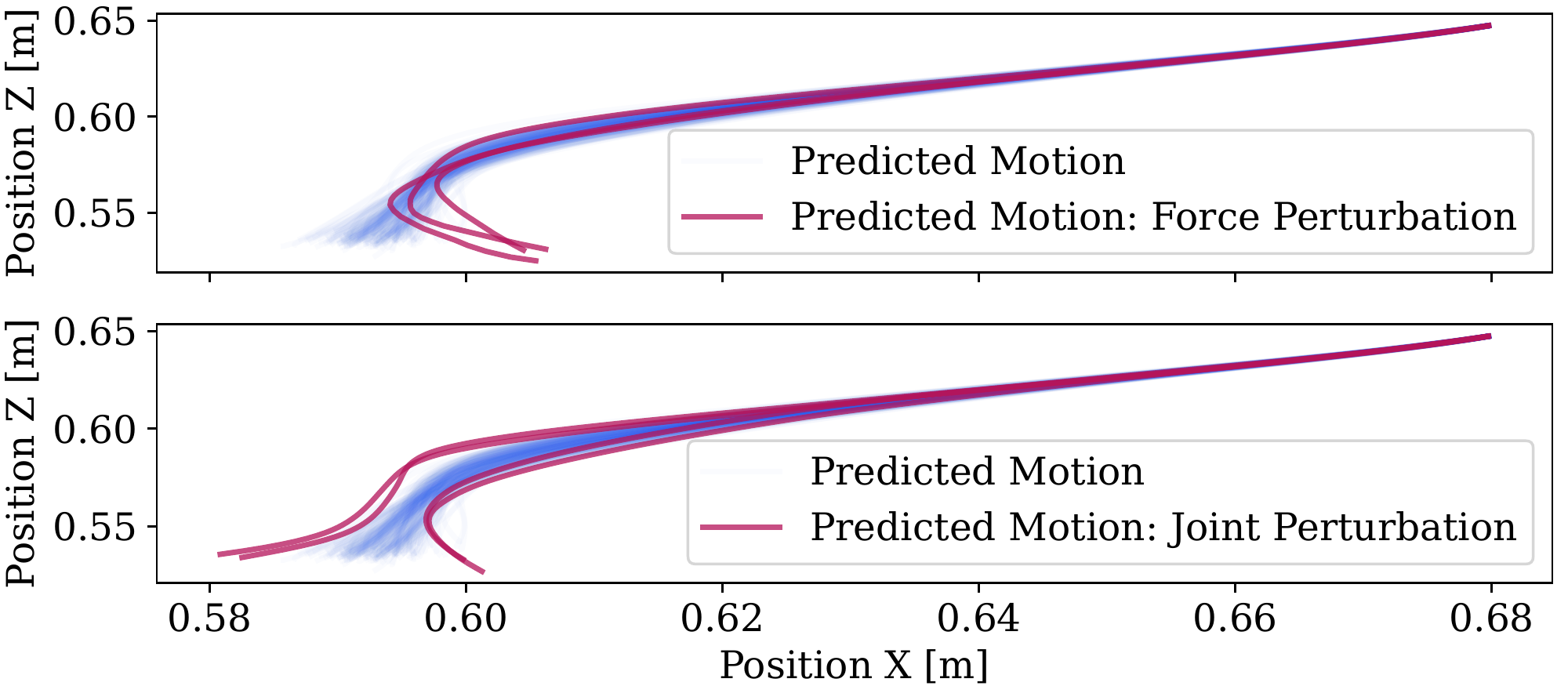}
     \caption{Generalization capabilities of our IntPrim model. Conditioned on the initial object pose, the model predicts similar trajectories over consecutive runs (blue), however, introducing disturbances to the F/T sensors (top) or joint position sensors (bottom) causes the model to recondition the motion (red), following the general motion trend, but unable to complete the task.}
     \label{fig:real_world_perturbation}
\end{figure}

While the model is able to recover from external perturbations, within reasonable limits, it is crucial that the recovery actions follow the demonstrated behavior. Figure~\ref{fig:real_world_perturbation} shows two scenarios in which either an external force, or an error in the joint sensors were introduced. In both cases, the model is able to adjust the motion accordingly (red lines) as compared to the demonstrated behaviors (blue lines), underlining our model's ability to adapt to these perturbations due to their connection learned in the EnBIP. 

\subsection{Baseline Comparison}

We also compare our model against a BC~\cite{pomerleau1991} and a ProMP~\cite{paraschos2013probabilistic} baseline in the real-world bimanual insertion task, as shown in Figure~\ref{fig:baselines}.
Similar to Bayesian Interaction Primitives, ProMPs are also derived from Dynamic Motor Primitives~\cite{ijspeert2013dynamical}. However, they do not include joint spatial and temporal reasoning. Instead, separate methods, such as Dynamic Time Warping (DTW)~\cite{ewerton2015learning} may have to be used for temporal reasoning. For the purposes of our comparison, we used ProMPs in conjunction with DTW which resulted in a success rate of 20\%. BC did not successfully complete any of the 30 attempts.
By comparison, BIC achieves a success rate of 90\%. These results highlight again the role of dynamic phase estimation in successful task completion.

\section{Conclusion}

In this paper, we introduced a framework for learning compliant bimanual policies from human demonstrations in contact-rich environments. Our approach combines the capabilities of admittance control and machine learning in order to enable efficient use of contacts with the environment while being able to adapt to a variety of external disturbances. We show that spatio-temporal inference plays a major role in creating safe, reliable and efficient control signals for physical interaction with objects and humans in a complex bimanual insertion task. Using the introduced methodology we achieve a success rate of $90\%$. Further, we have conducted a user study to identify optimal data collections interfaces. Our study shows that participants largely preferred kinesthetic teaching to teleoperation with a Space-Mouse; however, the multimodal data from the Space-Mouse provides crucial information for significantly improved task performance of the learned policy.\\

\noindent
\emph{Limitations:}
 While our approach is sample-efficient and can adapt to a range of perturbations, it is limited by the quality of the training data. The demonstrated behavior is assumed to be near-optimal in order to allow for precise phase estimation. Accordingly, the current system needs to be extended to better adjust for demonstrations from naive teachers~\cite{Chernova_2014}.

\bibliographystyle{IEEEtran}
\bibliography{IEEEabrv,references}

\begin{thebibliography}{10}
\providecommand{\url}[1]{#1}
\csname url@rmstyle\endcsname
\providecommand{\newblock}{\relax}
\providecommand{\bibinfo}[2]{#2}
\providecommand\BIBentrySTDinterwordspacing{\spaceskip=0pt\relax}
\providecommand\BIBentryALTinterwordstretchfactor{4}
\providecommand\BIBentryALTinterwordspacing{\spaceskip=\fontdimen2\font plus
\BIBentryALTinterwordstretchfactor\fontdimen3\font minus
  \fontdimen4\font\relax}
\providecommand\BIBforeignlanguage[2]{{%
\expandafter\ifx\csname l@#1\endcsname\relax
\typeout{** WARNING: IEEEtran.bst: No hyphenation pattern has been}%
\typeout{** loaded for the language `#1'. Using the pattern for}%
\typeout{** the default language instead.}%
\else
\language=\csname l@#1\endcsname
\fi
#2}}

\bibitem{billard2019}
A.~Billard and D.~Kragic, ``Trends and challenges in robot manipulation,''
  \emph{Science}, vol. 364, p. eaat8414, 06 2019.

\bibitem{kroemer2021review}
O.~Kroemer, S.~Niekum, and G.~D. Konidaris, ``A review of robot learning for
  manipulation: Challenges, representations, and algorithms,'' \emph{Journal of
  machine learning research}, vol.~22, no.~30, 2021.

\bibitem{koga1992}
Y.~Koga and J.-C. Latombe, ``Experiments in dual-arm manipulation planning,''
  in \emph{Proceedings 1992 IEEE International Conference on Robotics and
  Automation}, 1992, pp. 2238--2245 vol.3.

\bibitem{vahrenkamp2009humanoid}
N.~Vahrenkamp, D.~Berenson, T.~Asfour, J.~Kuffner, and R.~Dillmann, ``Humanoid
  motion planning for dual-arm manipulation and re-grasping tasks,'' in
  \emph{2009 IEEE/RSJ International Conference on Intelligent Robots and
  Systems}.\hskip 1em plus 0.5em minus 0.4em\relax IEEE, 2009, pp. 2464--2470.

\bibitem{cuong2018}
P.~Lertkultanon and Q.-C. Pham, ``A certified-complete bimanual manipulation
  planner,'' \emph{IEEE Transactions on Automation Science and Engineering},
  vol.~15, no.~3, pp. 1355--1368, 2018.

\bibitem{luck2017}
K.~S. Luck and H.~Ben~Amor, ``Extracting bimanual synergies with reinforcement
  learning,'' in \emph{2017 IEEE/RSJ International Conference on Intelligent
  Robots and Systems (IROS)}, 2017, pp. 4805--4812.

\bibitem{amadio2019}
F.~Amadio, A.~Colomé, and C.~Torras, ``Exploiting symmetries in reinforcement
  learning of bimanual robotic tasks,'' \emph{IEEE Robotics and Automation
  Letters}, vol.~4, no.~2, pp. 1838--1845, 2019.

\bibitem{lioutikov2016learning}
R.~Lioutikov, O.~Kroemer, G.~Maeda, and J.~Peters, ``Learning manipulation by
  sequencing motor primitives with a two-armed robot,'' in \emph{Intelligent
  Autonomous Systems 13}.\hskip 1em plus 0.5em minus 0.4em\relax Springer,
  2016, pp. 1601--1611.

\bibitem{pastor2011}
P.~Pastor, M.~Kalakrishnan, S.~Chitta, E.~Theodorou, and S.~Schaal, ``Skill
  learning and task outcome prediction for manipulation,'' in \emph{2011 IEEE
  International Conference on Robotics and Automation}, 2011.

\bibitem{ijspeert2013dynamical}
A.~J. Ijspeert, J.~Nakanishi, H.~Hoffmann, P.~Pastor, and S.~Schaal,
  ``Dynamical movement primitives: learning attractor models for motor
  behaviors,'' \emph{Neural computation}, vol.~25, no.~2, pp. 328--373, 2013.

\bibitem{billard2008handbook}
A.~Billard, S.~Calinon, R.~Dillmann, and S.~Schaal, ``Handbook of robotics
  chapter 59: robot programming by demonstration,'' \emph{Handbook of Robotics.
  Springer}, 2008.

\bibitem{hirzinger2004}
G.~Hirzinger, N.~Sporer, M.~Schedl, J.~Butterfaß, and M.~Grebenstein,
  ``Torque-controlled lightweight arms and articulated hands: Do we reach
  technological limits now?'' \emph{The International Journal of Robotics
  Research}, vol.~23, no. 4-5, pp. 331--340, 2004.

\bibitem{pomerleau1991}
D.~A. Pomerleau, ``Efficient training of artificial neural networks for
  autonomous navigation,'' \emph{Neural Computation}, vol.~3, no.~1, pp.
  88--97, 1991.

\bibitem{paraschos2013probabilistic}
A.~Paraschos, C.~Daniel, J.~R. Peters, and G.~Neumann, ``Probabilistic movement
  primitives,'' \emph{Advances in neural information processing systems},
  vol.~26, 2013.

\bibitem{amor2014interaction}
H.~B. Amor, G.~Neumann, S.~Kamthe, O.~Kroemer, and J.~Peters, ``Interaction
  primitives for human-robot cooperation tasks,'' in \emph{2014 IEEE
  international conference on robotics and automation (ICRA)}.\hskip 1em plus
  0.5em minus 0.4em\relax IEEE, 2014, pp. 2831--2837.

\bibitem{campbell2017bayesian}
J.~Campbell and H.~B. Amor, ``Bayesian interaction primitives: A slam approach
  to human-robot interaction,'' in \emph{Conference on Robot Learning}.\hskip
  1em plus 0.5em minus 0.4em\relax PMLR, 2017, pp. 379--387.

\bibitem{campbell2019probabilistic}
J.~Campbell, S.~Stepputtis, and H.~Ben~Amor, ``Probabilistic multimodal
  modeling for human-robot interaction tasks,'' in \emph{Robotics: Science and
  Systems}, 2019.

\bibitem{Roth2017}
M.~Roth, G.~Hendeby, C.~Fritsche, and F.~Gustafsson, ``The ensemble kalman
  filter: a signal processing perspective,'' \emph{{EURASIP} Journal on
  Advances in Signal Processing}, vol. 2017, no.~1, Aug. 2017.

\bibitem{thrun2005probabilistic}
S.~Thrun, W.~Burgard, and D.~Fox, \emph{Probabilistic robotics}.\hskip 1em plus
  0.5em minus 0.4em\relax Cambridge, Mass.: MIT Press, 2005.

\bibitem{todorov2012mujoco}
E.~Todorov, T.~Erez, and Y.~Tassa, ``Mujoco: A physics engine for model-based
  control,'' in \emph{2012 IEEE/RSJ International Conference on Intelligent
  Robots and Systems}.\hskip 1em plus 0.5em minus 0.4em\relax IEEE, 2012, pp.
  5026--5033.

\bibitem{nasa_tlx}
S.~Hart, ``Nasa-task load index (nasa-tlx); 20 years later,'' 2006.

\bibitem{nasa_tlx_original}
S.~G. Hart and L.~E. Staveland, ``Development of nasa-tlx (task load index):
  Results of empirical and theoretical research,'' in \emph{Human Mental
  Workload}, ser. Advances in Psychology, 1988.

\bibitem{ewerton2015learning}
M.~Ewerton, G.~Neumann, R.~Lioutikov, H.~B. Amor, J.~Peters, and G.~Maeda,
  ``Learning multiple collaborative tasks with a mixture of interaction
  primitives,'' in \emph{ICRA}.\hskip 1em plus 0.5em minus 0.4em\relax IEEE,
  2015, pp. 1535--1542.

\bibitem{Chernova_2014}
S.~Chernova and A.~L. Thomaz, \emph{Robot Learning from Human Teachers}.\hskip
  1em plus 0.5em minus 0.4em\relax Springer International Publishing, 2014.

\end{thebibliography}

\end{document}